\documentclass{article}

\usepackage{graphicx} 
\usepackage{subfigure} 

\usepackage{natbib}

\usepackage{algorithm}
\usepackage{algorithmic}

\usepackage{hyperref}

\usepackage[accepted]{icml2013}
\usepackage{amsmath}

\icmltitlerunning{PBODL : Parallel Bayesian Online Deep Learning for Click-Through Rate Prediction}

\begin{document} 

\twocolumn[
\icmltitle{PBODL : Parallel Bayesian Online Deep Learning for Click-Through Rate Prediction in Tencent Advertising System}

\icmlauthor{Xun Liu}{reubenliu@tencent.com}
\icmlauthor{Wei Xue}{weixue@tencent.com}
\icmlauthor{Lei Xiao}{shawnxiao@tencent.com}
\icmlauthor{Bo Zhang}{peanutzhang@tencent.com}


\vskip 0.3in
]

\begin{abstract}
We describe a parallel bayesian online deep learning framework (PBODL) for click-through rate (CTR) prediction within today's Tencent advertising system, which provides quick and accurate learning of user preferences. We first explain the framework with a deep probit regression model, which is trained with probabilistic back-propagation in the mode of assumed Gaussian density filtering. Then we extend the model family to a variety of bayesian online models with increasing feature embedding capabilities, such as Sparse-MLP, FM-MLP and FFM-MLP. Finally, we implement a parallel training system based on a stream computing infrastructure and parameter servers. Experiments with public available datasets and Tencent industrial datasets show that models within our framework perform better than several common online models, such as AdPredictor, FTRL-Proximal and MatchBox.  Online A/B test within Tencent advertising system further proves that our framework could achieve CTR and CPM lift by learning more quickly and accurately.
\end{abstract} 

\section{Introduction}
\label{Introduction}

Online advertising is a multi-billion dollar industry and is growing significantly each year. Just like in other online advertising settings, e.g., sponsored search, predicting ad click-through rates plays a central role in online advertising, since it impacts both user experience and profitability of the whole advertising system. 

\subsection{Tencent Advertising System}

In Tencent advertising platform, we choose ads from millions of candidates and serve them to hundreds of millions of users every day. In contrast to sponsored search, it is difficult for us to find out the instant needs of users, since we don't have search keywords by the time serving ads to them. As a result, advertising recommendations are usually made based on user's historical behaviors and context information.

Click-Through rate in advertising settings can be modeled as:

$ctr = P(click|user, ad, context)$

Features used in the conditional part are usually divided into three categories : 
\begin{itemize}
\item {\bf user features :} age, gender, interest, etc;
\item {\bf Ad features :} advertiser id, ad plan id, material id, ad industry, etc;
\item {\bf Context feature :} time, location, connection type, etc. 
\end{itemize}

The click-through rate prediction in tencent advertising platform faces many challenges : 

\begin{itemize}
\item {\bf New Ads :} Ad repository within our advertising system is updated frequently. 10\% to 20\% of the inventory could be replaced with new ads in each day. At the same time, advertisers are allowed to change their targeting rules during the campaign. Apparently, a daily or hourly updated model is not quick enough to support such kind of constant changes, and usually results in suboptimal revenue and advertisers' complaints according to our experience. 
\item {\bf Features Engineering :} As in all machine learning scenarios, features used dominate the performance of click-through rate prediction models. Unfortunately, feature engineering is a very time and resource consuming process. It demands more time and resource when you take more features into account, and could become the bottleneck of the whole modeling process.
\item {\bf Low latency :} The runtime latency spent on CTR prediction could not be large without negative impacts on user experience. In mobile advertising scenario, our latency budget for CTR prediction is about 10 ms. Considering the number of candidate ads, this is a very limited budget.
\item {\bf Big Data :} Click-through rate prediction is a massive-scale learning problem from all perspectives with no doubt. Billions of training samples are generated each day, leaving out the number of users and ads involved.
\end{itemize}

In the face of the above challenges, we set the following design goals for our solution: 

\begin{itemize}
\item {\bf Rapid Model Update :} We think rapid model update is one of the most effective ways to deal with constantly changing recommendation scenarios. Considering the scale of the learning problem, online learning (bayesian online learning in particular) paradigm  is chosen as a corner stone of our solution.
\item {\bf Nonlinear Model :} Linear models depends heavily on feature engineering, which is both time and resource consuming. Besides that, the size of a linear model explodes quickly while introducing more higher-order features. Inspired by the end-to-end training idea of deep learning, we choose deep nonlinear models as another corner stone of our solution.
\item {\bf Balance of Complexity and Latency :} Since the latency budget for CTR predictions per request is limited (10ms), deep models with dozens or hundreds of layers are impractical for real-life deployment. In fact, deep models with 3 ~ 5 hidden layers are proposed currently. In addition, each layer/operation involved should be designed and implemented as efficiently as possible (scale up).
\item {\bf Scale Out :} The solution should be able to scale out so as to support billions or even tens of billions of training samples per day.
\end{itemize}

\subsection{Related work}

Logistic regression with cross-features is an early well-known solution to CTR problem\cite{ctrnad,spad}. But it depends on complex features engineering and suffers curse of dimensionality. OWLQN\cite{owlqn} proposed by Microsoft can effectively pruning model and has been widely applied in many recommendation scenarios. 

In recent years, nonlinear models and online learning have got great attention in the field of CTR prediction. Factorization Machines (FM)\cite{fm,libfm} depends less on features engineering and is widely used in various competitions. A variant of FM called Field-aware Factorization Machines (FFM) has been used to win two click-through rate prediction competitions hosted by Criteo and Avazu\cite{ffm}. GBDT is another common solution in click-through rate prediction competitions and also widely used in industry. 

Inspired by deep learning\cite{cnn,rnn}, deep neural network is getting more popular in both competitions and industrial scenarios. But traditional deep neural networks cannot be directly applied to high-dimensional feature spaces, so many studies have focused on constructing better embedding layers. Sampling-based Neural Network (SNN) uses a regular embedding layer and Factorisation Machine supported Neural Network (FNN) initializes the embedding layer with the result of a pre-trained FM model\cite{nnctr}. Unlike FNN, Product-based Neural Network (PNN)\cite{pnn} can end-to-end learn the local dependencies which is similar to FM. Besides studis in embedding layers, Wide and Deep model\cite{wd} proposed by Google combines deep neural networks and linear models and significantly increases app acquisitions compared with wide-only and deep-only models in Google Play store. Convolutional neural networks (CNN) or Recurrent neural networks (RNN) are also tried with applications in CTR systems\cite{cnnctr,rnnctr}.

Although the models above have achieved state of art performance at their time, their experiments were mainly done on offline datasets with batch training. In real world online advertising scenarios, the system has to balance the accuracy of the model on history (training) data and the latency before pushing the model online. So many companies have tried online learning paradigm. AdPredictor\cite{adpredictor} proposed by Microsoft and FTRL\cite{ftrl} proposed by Google are two of the most famous models. But these two models are both linear models and suffer features engineering related costs. Facebook introduces a hybrid model which combines decision trees with online linear model and the new model outperforms either of these methods on their own by over 3\%\cite{gbdtlr}. MatchBox\cite{matchbox} proposed by Microsoft is an online matrix factorization model but it's hard to port its learning algorithm to more complex models like DNN.

In this paper, we describe a parallel bayesian online deep learning framework used for click through rate prediction in Tencent advertising system. Section 2 introduces the framework with a bayesian deep probit model trained with probabilistic backpropagation in the mode of assumed Gaussian density filtering\cite{pbp,ep}, then extends the model famility with some novel online deep models and presents a parallel model updating framework\cite{svb,ep}. Section 3 compares these models with common models such as AdPredictor, FTRL, FM and MLP in both offline datasets and online experiments. Section 4 gives some notes when we apply this framework. Finally, Section 5 presents conclusions and future work.

The main contributions of the paper are:
\begin{itemize}
\item A general parallel bayesian online deep learning framework sutiable for a variety of bayesian online models. 
\item Some novel bayesian online deep model with efficient training and predicting operations.
\item We applied the framework and these bayesian online models in the Tencent Advertising system and get significant improvements over commonly used models.
\end{itemize}

\section{Parallel Bayesian Online Deep Learning Framework}
\label{PBODLF}

In this section we present a bayesian online learning framework. In this framework, we can implement variety of models whose parameters can be real-time updated. In Section 2.1 we describe the deep probit model for CTR and how to inference on it. In Section 2.2 we describe several novel deep models in this framework. In Section 2.3 we describe how to parallelly update parameters in an easy way.

\begin{figure}[ht]
\vskip 0.2in
\begin{center}
\centerline{\includegraphics[width=\columnwidth]{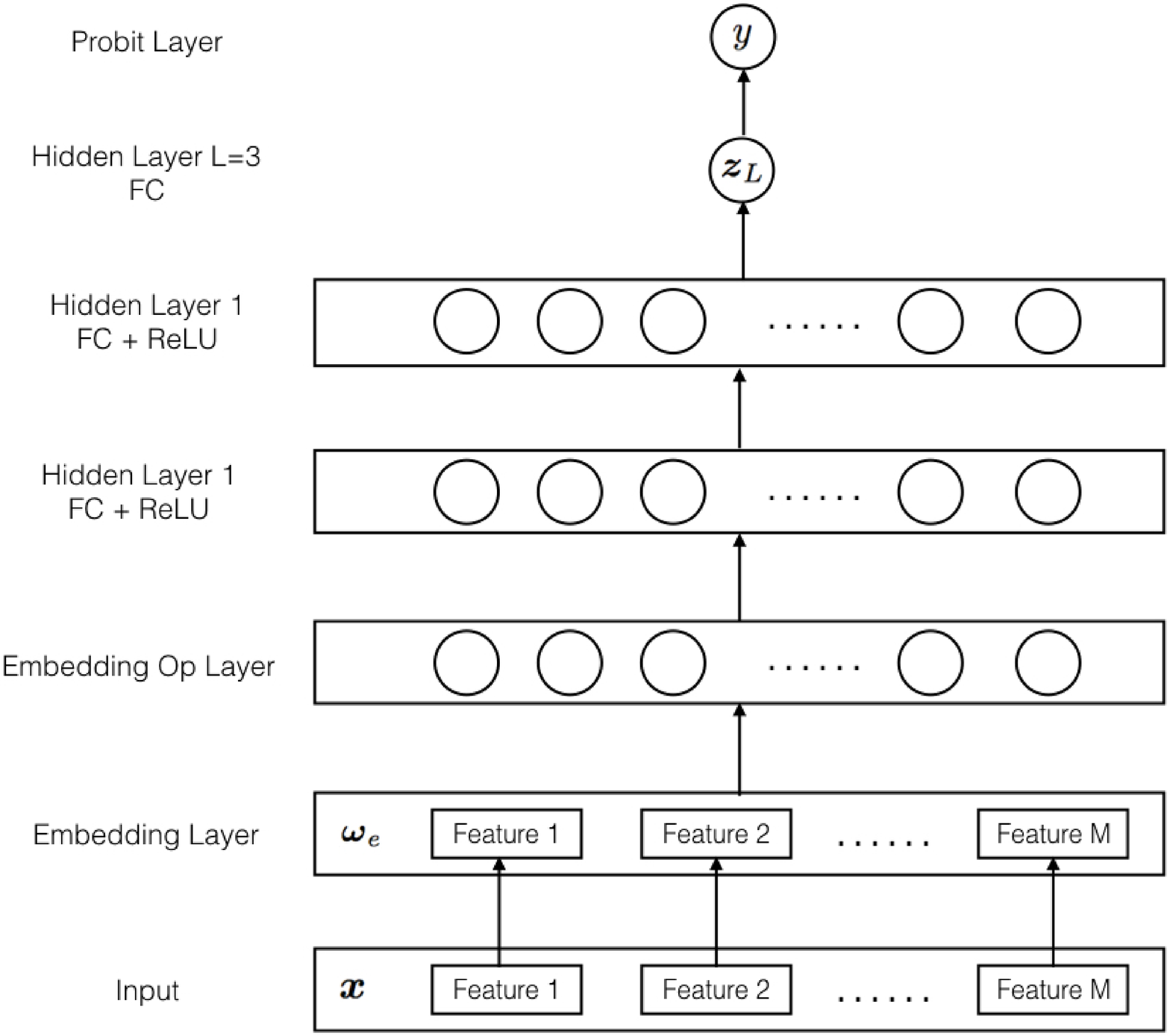}}
\caption{Bayesian Deep Probit Model}
\label{fig_bdpm}
\end{center}
\vskip -0.2in
\end{figure} 

\subsection{Bayesian Deep Probit Model}
\label{BDPM}

The deep probit model is a feed-forward neural network with a probit link function like Figure \ref{fig_bdpm}.

Given a data point$(\boldsymbol{x}, y)$, made up of a input feature vector $\boldsymbol{x}$ and a label variable $y$, we assume that $y$ is obtained as : 
{\setlength\abovedisplayskip{1pt}
\setlength\belowdisplayskip{1pt}
\begin{equation}
\begin{split}
y = sign(f(\boldsymbol{x}, \boldsymbol{\omega}) + \epsilon)
\end{split}
\end{equation}}
where $f(\boldsymbol{x}, \boldsymbol{\omega})$ is the output of the feed-forward neural network with weights given by $\boldsymbol{\omega}$ and $\boldsymbol{x}$, and $\epsilon$ is an additive noise, where $\epsilon \sim N(0, 1)$.

Because of sparse input features, the first layer of the model is a embedding layer, where $\boldsymbol{\omega}_e$ is its weight and $\boldsymbol{z}_e$ is its output. For get better embedding of sparse input features, we add a common embedding op layer which is similar to inner product layer in PNN. In general, this layer is parameterless. Considered the amount of calculation, we introduce three effective embedding op layer. Then the output of embedding op layer is used as the input of a common deep neural model with multiple hidden layers. The number of hidden layers is $L$. There are $V_l$ hidden units in layer $l\in{[1, L]}$ and $\boldsymbol{\omega}_l$ is the weight matrices between hidden layer ${l-1}$ and hidden layer $l$. We denote the output of layer $l$ by $\boldsymbol{z}_l$ and the input of layer $l$ by $\boldsymbol{a}_l=\boldsymbol{\omega}_l\boldsymbol{z}_{l-1}/\sqrt{V_{l-1} + 1}$. Specially, $\boldsymbol{z}_0$ is the output of embedding op layer and $\boldsymbol{z}_L = f(\boldsymbol{x}, \boldsymbol{\omega})$. The activation functions for each hidden layer are rectified linear units (ReLUs).

We have described the general structure of our CTR model. Then we need to know how to inference on the model. Given input feature vector $\boldsymbol{x}$ and $\boldsymbol{w}$, the CTR can be written as :
{\setlength\abovedisplayskip{1pt}
\setlength\belowdisplayskip{1pt}
\begin{equation}
\begin{split}
p(y|\boldsymbol{x}, \boldsymbol{w}) &= \Phi(y\boldsymbol{z}_L)
\end{split}
\end{equation}}

To complete our probabilistic model, we specify a Gaussian prior distribution for each of $\boldsymbol{w}$. In particular,
{\setlength\abovedisplayskip{1pt}
\setlength\belowdisplayskip{1pt}
\begin{equation}
\begin{split}
p(\boldsymbol{w}) &= p(\boldsymbol{\omega}_e)\prod_{l=1}^{L}{p(\boldsymbol{\omega}_l)} \\
				&= \prod_{i=0}^{N_e}N(\omega_{e,i}|m_e, v_e)\prod_{l=1}^{L}\prod_{j=1}^{N_l}{N(\omega_{l,j}|m_l, v_l)}
\end{split}
\end{equation}}
where $m_e,m_l$ is prior mean and $v_e,v_l$ is prior variance.

Considered the prior distribution, predictive distribution can be derived as
{\setlength\abovedisplayskip{1pt}
\setlength\belowdisplayskip{1pt}
\begin{equation}
\begin{split}
p(y|\boldsymbol{x}) &= E_{p(\boldsymbol{\omega})}[\Phi(y\boldsymbol{z}_L)]
\end{split}
\end{equation}}

Given $p(y|\boldsymbol{x}, \boldsymbol{w})$ and the prior $p(\boldsymbol{w})$, the posterior distribution for the parameters $\omega$ can then be obtained by applying Bayes’ rule:
{\setlength\abovedisplayskip{1pt}
\setlength\belowdisplayskip{1pt}
\begin{equation}
\begin{split}
p(\boldsymbol{\omega}|\boldsymbol{x}, y) \propto p(y|\boldsymbol{x}, \boldsymbol{w}) \cdot p(\boldsymbol{w})
\end{split}
\end{equation}}
Both the exact predictive distribution and exact posterior over weights can't be calculated in simple closed form. So we use probabilistic backpropagation(PBP)\cite{pbp} in the mode of assumed Gaussian density filtering to approximate them. PBP is a scalable method for learning Bayesian neural networks. Similar to classical backpropagation, PBP works by computing a forward propagation of probabilities through the network and then doing a backward computation of gradients. First, PBP propagates
distributions forward through the network and approximates each new distribution with a Gaussian when necessary. Then PBP computes the gradients of weights and update each of $\boldsymbol{\omega}$ with following rules\cite{ep,pbp}:
{\setlength\abovedisplayskip{1pt}
\setlength\belowdisplayskip{1pt}
\begin{align}
m^{new} &= m + v\frac{\partial{logZ}}{\partial{m}} \\
v^{new} &= v - v^2[(\frac{\partial{logZ}}{\partial{m}})^2 - 2\frac{\partial{logZ}}{\partial{v}}]
\label{eql:adf}
\end{align}}
where $Z$ is the the normalization constant :
{\setlength\abovedisplayskip{1pt}
\setlength\belowdisplayskip{1pt}
\begin{align}
Z = \int_{\boldsymbol{w}}{p(y|\boldsymbol{x}, \boldsymbol{w}) \cdot p(\boldsymbol{w})} = p(y|\boldsymbol{x})
\label{z_1}
\end{align}}

When predicting with model, we only need to do a forward computation. Because PBP will approximates each new distribution with a Gaussian when necessary. So $\boldsymbol{z}_L$, the output of last hidden layer will be be approximated as a Gaussian variable with mean $m^{\boldsymbol{z}_L}$ and variance $v^{\boldsymbol{z}_L}$. So the predictive distribution can be approximated as:
{\setlength\abovedisplayskip{1pt}
\setlength\belowdisplayskip{1pt}
\begin{equation}
\begin{split}
p(y|\boldsymbol{x}) &\approx \Phi(y\frac{m^{\boldsymbol{z}_L}}{\sqrt{v^{\boldsymbol{z}_L} + 1}})
\label{p_y_x_2}b
\end{split}
\end{equation}}

To update the weights of model, we need to compute the normalization constant $Z$. With  equations \eqref{z_1} and \eqref{p_y_x_2}, we can approximate $Z$ as :
{\setlength\abovedisplayskip{1pt}
\setlength\belowdisplayskip{1pt}
\begin{equation}
\begin{split}
Z = p(y|\boldsymbol{x}) &\approx \Phi(y\frac{m^{\boldsymbol{z}_L}}{\sqrt{v^{\boldsymbol{z}_L} + 1}})
\end{split}
\end{equation}}

However, it remains to compute the mean and variance parameters $m^{\boldsymbol{z}_L}$ and $v^{\boldsymbol{z}_L}$ through the network. In embedding layer, $\boldsymbol{x}_i$, the $i$-th element of $\boldsymbol{x}$ is transformed to a dense value vector $\boldsymbol{\omega}_{e,\boldsymbol{x}_i}$.
Then all dense value vectors are concated as $\boldsymbol{z}_e$.
{\setlength\abovedisplayskip{1pt}
\setlength\belowdisplayskip{1pt}
\begin{equation}
\begin{split}
\boldsymbol{z}_e &= (\boldsymbol{\omega}_{e,\boldsymbol{x}_0}, \boldsymbol{\omega}_{e,\boldsymbol{x}_1}, ..., \boldsymbol{\omega}_{e,\boldsymbol{x}_M}) \\
\boldsymbol{m}^{\boldsymbol{z}_e} &= (\boldsymbol{m}^{\boldsymbol{\omega}_{e,\boldsymbol{x}_0}}, \boldsymbol{m}^{\boldsymbol{\omega}_{e,\boldsymbol{x}_1}}, ..., \boldsymbol{m}^{\boldsymbol{\omega}_{e,\boldsymbol{x}_M}}) \\
\boldsymbol{v}^{\boldsymbol{z}_e} &= (\boldsymbol{v}^{\boldsymbol{\omega}_{e,\boldsymbol{x}_0}}, \boldsymbol{v}^{\boldsymbol{\omega}_{e,\boldsymbol{x}_1}}, ..., \boldsymbol{v}^{\boldsymbol{\omega}_{e,\boldsymbol{x}_M}}) 
\end{split}
\end{equation}}

In embedding op layer, we can design a variety of operations to get better embedding. The simplest operation is copying input to output :
{\setlength\abovedisplayskip{1pt}
\setlength\belowdisplayskip{1pt}
\begin{align}
\boldsymbol{z}_0 = \boldsymbol{z}_e, \qquad
\boldsymbol{m}^{\boldsymbol{z}_0} = \boldsymbol{m}^{\boldsymbol{z}_e}, \qquad
\boldsymbol{v}^{\boldsymbol{z}_0} = \boldsymbol{v}^{\boldsymbol{z}_e}
\end{align}}
More operations will be introduced in Section \ref{model}.

In $l$-th hidden layer, using moment match, the mean and variance of  $\boldsymbol{a}_l$ can be approximated as :
{\setlength\abovedisplayskip{1pt}
\setlength\belowdisplayskip{1pt}
\begin{equation}
\begin{split}
\boldsymbol{m}^{\boldsymbol{a}_l} &= \boldsymbol{m}^{\boldsymbol{\omega}_l}\boldsymbol{m}^{\boldsymbol{z}_{l-1}}/\sqrt{V_l+1} \\
\boldsymbol{v}^{\boldsymbol{a}_l} &= [\boldsymbol{v}^{\boldsymbol{\omega}_l}\boldsymbol{v}^{\boldsymbol{z}_{l-1}}+(\boldsymbol{m}^{\boldsymbol{\omega}_l} \circ \boldsymbol{m}^{\boldsymbol{\omega}_l}) \boldsymbol{v}^{\boldsymbol{z}_{l-1}} \\
& + \boldsymbol{v}^{\boldsymbol{\omega}_l} (\boldsymbol{m}^{\boldsymbol{z}_{l-1}} \circ \boldsymbol{m}^{\boldsymbol{z}_{l-1}})]/(V_l+1)
\end{split}
\end{equation}}
Let $\boldsymbol{z}_l = max(0, \boldsymbol{a}_l)$, the mean and variance of the $i$-th element of $\boldsymbol{z}_l$ can be approximated as :
{\setlength\abovedisplayskip{1pt}
\setlength\belowdisplayskip{1pt}
\begin{equation}
\begin{split}
\boldsymbol{m}_i^{\boldsymbol{z}_l} &= \Phi(\alpha_i)v'_i \\
\boldsymbol{v}_i^{\boldsymbol{z}_l} &= \Phi(\alpha_i)\boldsymbol{v}_i^{\boldsymbol{z}_l}(1-\gamma_i(\gamma_i + \alpha_i)) \\
&+ \boldsymbol{m}_i^{\boldsymbol{z}_l}\Phi(-\alpha_i)v'_i
\end{split}
\end{equation}}
where
{\setlength\abovedisplayskip{1pt}
\setlength\belowdisplayskip{1pt}
\begin{small}
\begin{align}
v'_i = \boldsymbol{m}_i^{\boldsymbol{a}_l} + \sqrt{\boldsymbol{v}_i^{\boldsymbol{a}_l}}\gamma_i, \quad
\alpha_i = \frac{\boldsymbol{m}_i^{\boldsymbol{a}_l}}{\sqrt{\boldsymbol{v}_i^{\boldsymbol{a}_l}}}, \quad
\gamma_i = \frac{\phi(-\alpha_i)}{\Phi(\alpha_i)}
\end{align}
\end{small}}

Finally, we can get the mean and variance of $\boldsymbol{z}_L$.

\subsection{Embedding Operation Layer}
\label{model}
In \ref{BDPM}, we have described a deep probit model with a simplest embedding op layer and how to online learn it and predict with it. In our practice, we find that we can get better performance in online experiment by using better embedding operation layer such as FM and FFM. Here we introduce three novel embedding operations : DimensionAwareSum, FM and FFM. Corresponding to the three operations, we propose three novel models called Sparse-MLP, FM-MLP and FFM-MLP.

\subsubsection{DimensionAwareSum Layer}
The DimensionAwareSum layer is used in Sparse-MLP as Embedding Operation Layer. When using DimensionAwareSumLayer, we should encode every sparse input feature into dense value vector with fixed length $K$. So if the nonzero of input $\boldsymbol{x}$ is $M$, $\boldsymbol{z}_e$ is a $M\times{K}$ matrices. In general, 
{\setlength\abovedisplayskip{1pt}
\setlength\belowdisplayskip{1pt}
\begin{equation}
\begin{split}
\boldsymbol{z}_{e,i,j} = \boldsymbol{\omega}_{e,\boldsymbol{x}_i, j} \quad i=1...M, j=1...K
\end{split}
\end{equation}}
where $\boldsymbol{\omega}_{e,\boldsymbol{x}_i, j}$ is the j-th element of dense embedding vector of $\boldsymbol{x}_i$.

The output $\boldsymbol{z}_0$ is a $K\times{1}$ matrices and satisfy :
{\setlength\abovedisplayskip{1pt}
\setlength\belowdisplayskip{1pt}
\begin{equation}
\begin{split}
\boldsymbol{z}_{0,j} = \sum_{i=1}^{M}{\boldsymbol{z}_{e,i,j}} \quad i=1...M, j=1...K
\end{split}
\end{equation}}
The mean and variance of $\boldsymbol{z}_{0,j}$ can be computed as :
{\setlength\abovedisplayskip{1pt}
\setlength\belowdisplayskip{1pt}
\begin{equation}
\begin{split}
m^{\boldsymbol{z}_0}_j &= \sum_{i=1}^{M}{m^{\boldsymbol{z}_{e}}_{i,j}} \quad i=1...M, j=1...K \\
v^{\boldsymbol{z}_0}_j &= \sum_{i=1}^{M}{v^{\boldsymbol{z}_{e}}_{i,j}} \quad i=1...M, j=1...K
\end{split}
\end{equation}}

\subsubsection{FM Layer}
The FM layer is used in FM-MLP as Embedding Operation Layer. Similar to PNN, we consider product relationship with input. But unlike IPNN, the dimension of $\boldsymbol{z}_0$ in FM Layer is $K$ where the result of inner product layer is a $M\times{M}$ matrices. In general, $M\times{M} \gg K$. In addition, $\boldsymbol{z}_0$ isn't non-negative as IPNN and OPNN.
The $k$-th element of $\boldsymbol{z}_0$ satisfy :
{\setlength\abovedisplayskip{1pt}
\setlength\belowdisplayskip{1pt}
\begin{equation}
\begin{split}
\boldsymbol{z}_{0,k} = \sum_{i=1}^{M}\sum_{j=i+1}^{M}{\boldsymbol{z}_{e,i,k}\boldsymbol{z}_{e,j,k}} \quad i, j=1...M, k=1...K
\end{split}
\end{equation}}
Use Moment Match for $\boldsymbol{z}_{e,i,k}\boldsymbol{z}_{e,j,k}$ and similar trick in FM, the mean and variance of $\boldsymbol{z}_{0,k}$ can be computed in $O(M)$ times.
{\setlength\abovedisplayskip{1pt}
\setlength\belowdisplayskip{1pt}
\begin{equation}
\begin{split}
m^{\boldsymbol{z}_0}_k &= \frac{1}{2}[(\sum_{i=1}^{M}{m^{\boldsymbol{z}_{e}}_{i,k}})^2 - \sum_{i=1}^{M}{(m^{\boldsymbol{z}_{e}}_{i,k})^2}] \\
v^{\boldsymbol{z}_0}_k &= \frac{1}{2}[(\sum_{i=1}^{M}{sm^{\boldsymbol{z}_{e}}_{i,k}})^2 - \sum_{i=1}^{M}{(sm^{\boldsymbol{z}_{e}}_{i,k})^2}] \\
&-\frac{1}{2}[(\sum_{i=1}^{M}{(m^{\boldsymbol{z}_{e}}_{i,k})^2})^2 - \sum_{i=1}^{M}{(m^{\boldsymbol{z}_{e}}_{i,k})^4}]
\end{split}
\end{equation}}
where $sm^{\boldsymbol{z}_{e}}_{i,k}$ is the second moment of $\boldsymbol{z}_{e,i,k}$ :
{\setlength\abovedisplayskip{1pt}
\setlength\belowdisplayskip{1pt}
\begin{equation}
\begin{split}
sm^{\boldsymbol{z}_{e}}_{i,k} = (m^{\boldsymbol{z}_{e}}_{i,k})^2 + v^{\boldsymbol{z}_{e}}_{i,k}
\end{split}
\end{equation}}

\subsubsection{FFM Layer}
The FFM layer is used in FFM-MLP as Embedding Operation Layer. FFM is a famous extension of FM and performance better than FM. In FFM layer, The $k$-th element of $\boldsymbol{z}_0$ satisfy :
{\setlength\abovedisplayskip{1pt}
\setlength\belowdisplayskip{1pt}
\begin{equation}
\begin{split}
\boldsymbol{z}_{0,k} = \sum_{i=1}^{M}\sum_{j=i+1}^{M}{\boldsymbol{z}_{e,i,f_j,k}\boldsymbol{z}_{e,f_i,j,k}} \\
i, j=1...M, \quad k=1...K, \quad 1 \leq f_i,f_j \leq F
\end{split}
\end{equation}}
where $F$ is the filed num and $\boldsymbol{z}_e$ is a $F\times{K}\times{M}$ matrices. Unlike DimensionAwareSum and FM Layer, the length of embedding vector of $\boldsymbol{x}_i$ is $F\times{K}$ and $f_i$ is the field that $\boldsymbol{x}_i$ belongs to.

Unfortunately, FFM can't reduce the amount of calculation as same as FM. The complexity of FFM is $O(KM^2)$ while that of FM is $O(KM)$. But when $F \ll M$, we can compute FFM in time $O(FKM + KF^2)$. In actual system such as tencent advertising, $F$ is usually small e.g 2 or 3 for reducing computation, saving memory and avoiding overfitting.
{\setlength\abovedisplayskip{1pt}
\setlength\belowdisplayskip{1pt}
\begin{equation}
\begin{split}
m^{\boldsymbol{z}_0}_k &= \sum_{f=1}^{F}{m^{A_f}} + \sum_{f_i=1}^{F}\sum_{f_j=f_i+1}^{F}{m^{B_{f_i,f_j}}} \\
v^{\boldsymbol{z}_0}_k &= \sum_{f=1}^{F}{v^{A_f}} + \sum_{f_i=1}^{F}\sum_{f_j=f_i+1}^{F}{v^{B_{f_i,f_j}}} \\
\end{split}
\end{equation}}
where
{\setlength\abovedisplayskip{1pt}
\setlength\belowdisplayskip{1pt}
\begin{equation}
\begin{split}
m^{A_f} &= \frac{1}{2}[(\sum_{i=1}^{M}{m^{\boldsymbol{z}_{e}}_{i,f,k}})^2 - \sum_{i=1}^{M}{(m^{\boldsymbol{z}_{e}}_{i,f,k})^2}] \\
v^{A_f} &= \frac{1}{2}[(\sum_{i=1}^{M}{sm^{\boldsymbol{z}_{e}}_{i,f,k}})^2 - \sum_{i=1}^{M}{(sm^{\boldsymbol{z}_{e}}_{i,f,k})^2}] \\
&-\frac{1}{2}[(\sum_{i=1}^{M}{(m^{\boldsymbol{z}_{e}}_{i,f,k})^2})^2 - \sum_{i=1}^{M}{(m^{\boldsymbol{z}_{e}}_{i,f,k})^4}] \\
m^{B_{f_i,f_j}} &= (\sum_{i=1}^{M}{m^{\boldsymbol{z}_{e}}_{i,f_i,k}})(\sum_{i=1}^{M}{m^{\boldsymbol{z}_{e}}_{i,f_j,k}}) \\
v^{B_{f_i,f_j}} &= (\sum_{i=1}^{M}{m^{\boldsymbol{z}_{e}}_{i,f_i,k}})^2(\sum_{i=1}^{M}{v^{\boldsymbol{z}_{e}}_{i,f_j,k}}) \\
& + (\sum_{i=1}^{M}{m^{\boldsymbol{z}_{e}}_{i,f_j,k}})^2(\sum_{i=1}^{M}{v^{\boldsymbol{z}_{e}}_{i,f_i,k}}) \\
& + (\sum_{i=1}^{M}{v^{\boldsymbol{z}_{e}}_{i,f_i,k}})(\sum_{i=1}^{M}{v^{\boldsymbol{z}_{e}}_{i,f_j,k}})
\end{split}
\end{equation}}
and $sm^{\boldsymbol{z}_{e}}_{i,f,k}$ is the second moment of $\boldsymbol{z}_{e,i,f,k}$ :
{\setlength\abovedisplayskip{1pt}
\setlength\belowdisplayskip{1pt}
\begin{equation}
\begin{split}
sm^{\boldsymbol{z}_{e}}_{i,f,k} = (m^{\boldsymbol{z}_{e}}_{i,f,k})^2 + v^{\boldsymbol{z}_{e}}_{i,f,k}
\end{split}
\end{equation}}

\subsection{Parallel Training}

Every day there are ten of billions of samples in tencent advertising system. The CTR model must be trained in parallel mode. Here we introduce a simple and effective parallel update framework for our Bayesian online models. 
\begin{figure}[!htbp]
\vskip 0.2in
\begin{center}
\centerline{\includegraphics[width=\columnwidth]{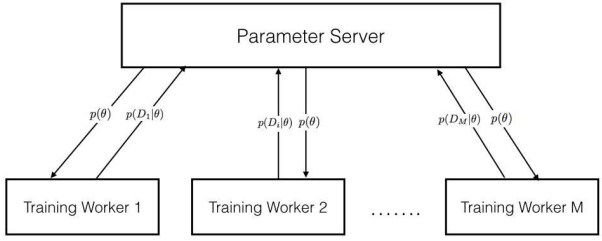}}
\caption{Parallel Training Framework}
\label{fig_ps}
\end{center}
\vskip -0.2in
\end{figure} 

We store global parameters in parameter server and use multiple workers in storm for training data. In time $T$, assumed the parameters in parameter server as prior $P(\theta)$ and minibatch data in $i$-th worker as $D_i$, then the posterior can be obtained by applying Bayes’ rule:
{\setlength\abovedisplayskip{1pt}
\setlength\belowdisplayskip{1pt}
\begin{equation}
\begin{split}
p(\theta|D_1,D_2,...,D_M) \propto p(\theta)\prod_{i=1}^Mp(D_i|\theta)
\end{split}
\end{equation}}

Because $p(\theta|D_1,D_2,...,D_M)$ and $p(\theta)$ are Gaussian, the likelihood $p(D_i|\theta)$ can be approximated as Gaussian. In $i$-th worker, we can first compute the $p(\theta|D_i)$ using PBP. The $p(D_i|\theta)$ can be obtained by equation \ref{msg} and then is updated to parameter server.
{\setlength\abovedisplayskip{1pt}
\setlength\belowdisplayskip{1pt}
\begin{equation}
\begin{split}
p(D_i|\theta) \propto p(\theta|D_i)/p(\theta)
\label{msg}
\end{split}
\end{equation}}

In the case of Gaussian, the flow of training worker and parameter server is as algorithm \ref{alg:worker} and algorithm \ref{alg:server}.

\begin{algorithm}[ht]
   \caption{Flow of Train Worker}
   \label{alg:worker}
\begin{algorithmic}
   \REPEAT
   \STATE {\bfseries Input:} minibatch data $D_t$ and minibatch size $N$
   \STATE Initialize $p(\boldsymbol{\omega})$ from parameter server.
   \FOR{$i=1$ {\bfseries to} $N$}
   \STATE Compute every $m^{new}_{\omega_k}$ and $v^{new}_{\omega_k}$ in $\boldsymbol{\omega}$ using equation \ref{eql:adf}
   \STATE Add the likelihood of $D_{t,i}$ using follow equation:\\
   \quad \quad $\frac{m^{looklike}_{\omega_k}}{v^{looklike}_{\omega_k}} += \frac{m^{new}_{\omega_k}}{v^{new}_{\omega_k}} - \frac{m_{\omega_k}}{v_{\omega_k}}$ \\ 
   \quad \quad $\frac{1}{v^{looklike}_{\omega_k}} += \frac{1}{v^{new}_{\omega_k}} - \frac{1}{v_{\omega_k}}$
   \STATE Update local parameter : $m_{\omega_k} = m^{new}_{\omega_k}, v_{\omega_k} = v^{new}_{\omega_k}$
   \ENDFOR
   \STATE Update $\{m^{looklike}_{\omega_k}, v^{looklike}_{\omega_k}\}$ as the likelihood of $D_t$ to parameter server
   \UNTIL
\end{algorithmic}
\end{algorithm}

\begin{algorithm}[ht]
   \caption{Update Flow of Parameter Server}
   \label{alg:server}
\begin{algorithmic}
   \REPEAT
   \STATE {\bfseries Input:} $\{\omega_k, m^{looklike}_k, v^{looklike}_k\}$, size of msg K
   \FOR{$k=1$ {\bfseries to} $K$}
   \STATE Get old value of $\omega_k$ : $m^{old}_k, v^{old}_k$
   \STATE Compute new value of $\omega_k$ using follow equation: 
   \STATE \quad $\frac{m^{new}_k}{v^{new}_k} = \frac{m^{old}_k}{v^{old}_k} + \frac{m^{looklike}_k}{v^{looklike}_k}$\\
   \STATE \quad $\frac{1}{v^{new}_k} = \frac{1}{v^{old}_k} + \frac{1}{v^{looklike}_k}$
   \STATE Update value of $\omega_k$ with $m^{new}_k, v^{new}_k$
   \ENDFOR
   \UNTIL
\end{algorithmic}
\end{algorithm}

Above framework can deal with big data in the real world. We have applied it into Tencent advertising system and got a steady performance.

\section{Experiment Result}
\label{Experiment Result}

In this section, we present our experiments in detail, including offline experiments and online experiments. In offline experiments, we compare our several online models with previous online models such as AdPredictor, FTRL and MatchBox. We compare them in a Avazu dataset and our internal dataset. In online experiments, we compare our online models with DNN with batch training.

\subsection{Offline}

\subsubsection{Avazu Dataset}

Avazu Dataset\footnote{https://www.kaggle.com/c/avazu-ctr-prediction} is from a competition of kaggle in 2014. For this competition, Avazu has provided a dataset with label of 10 days. We use the previous 9 days of data for training and the last 1 days of data for test. Unlike competition, we take more attention to the model rather than features. So we simply use the original 22 features besides id and device ip. We compare our several models with AdPredictor and FTRL without feature engineering and MatchBox. All models only train data once in chronological order. The result is shown in Table \ref{avazu_data}. From the result, we can find that our models are significantly better than AdPredictor and FTRL without feature engineering. Compared with MatchBox, our models also perform better on both AUC and Loss.

\begin{table}[ht]
\caption{AUC \& Loss in Avazu dataset.}
\label{avazu_data}
\vskip 0.15in
\begin{center}
\begin{small}
\begin{sc}
\begin{tabular}{lcccr}
\hline
\abovespace\belowspace
Model & AUC & LOSS \\
\hline
AdPredictor &  0.7375 & 0.4019 \\
FTRL & 0.7357 & 0.4030 \\
MatchBox    & 0.7426 & 0.4001 \\
Sprase-MLP(Online)  & 0.7489 & 0.3987 \\
FM-MLP(Online)      & 0.7489 & 0.3972 \\
FFM-MLP(Online)     & 0.7498 & 0.3981\\
\hline
\end{tabular}
\end{sc}
\end{small}
\end{center}
\vskip -0.1in
\end{table}

\subsubsection{Internal DataSet}

In this section, we use our internal dataset of tecent advertising. We use ten days of data to train and one day of data to test. The training dataset comprises 800M instances with 40 categorys of features. The result is shown in Table \ref{i-data}. We can find that our models perform better than other online models without feature engineering.

\begin{table}[ht]
\caption{AUC \& Loss in tencent internal dataset of advertising.}
\label{i-data}
\vskip 0.15in
\begin{center}
\begin{small}
\begin{sc}
\begin{tabular}{lcccr}
\hline
\abovespace\belowspace
Model & AUC & LOSS \\
\hline
AdPredictor &  0.7359 & 0.1210 \\
FTRL & 0.7365 & 0.1212 \\
MatchBox  & 0.7476 & 0.1198 \\
Sparse-MLP(Online)  & 0.7510 & 0.1192 \\
FM-MLP(Online)      & 0.7519 & 0.1191\\
FFM-MLP(Online)     & 0.7529 & 0.1190 \\
\hline
\end{tabular}
\end{sc}
\end{small}
\end{center}
\vskip -0.1in
\end{table}

\subsection{Online}

However, offline evaluation can only be used as reference and it isn't easy to compare online models with batch models in offline datasets. Online environment is more complex, so we generally using A/B test to evaluate the real effect of the models. We have applid our framework and serveral models into tencent advertising system and got a good online effect. Because of the difficulty of feature engineering, we only compare our models with FM and DNN using batch training which has achieved the best effect. The online effect is shown in Table \ref{online}. Our online models are better than DNN. An important reason is that our online models can update every 15 minutes while DNN can only update every several hours. So the online models can perform much better in new ads, just as Tabel \ref{newads}. Meanwhile, the percentage of new Ads using online models is higher than that using batch models, so it can get long term profit to use our bayesian online models. In addition, our online models also achieve cpm lift comparable with ctr lift.

\begin{table}[ht]
\caption{CTR Lift in online experiment}
\label{online}
\vskip 0.15in
\begin{center}
\begin{small}
\begin{sc}
\begin{tabular}{lcccr}
\hline
\abovespace\belowspace
Model & Lift \\
\hline
FM(Batch)    &  BaseLine \\
DNN(Batch)    & +2.2\% \\
FM-MLP(Online)      &  +4.1\% \\
FFM-MLP(Online)      & +5.6\% \\
\hline
\end{tabular}
\end{sc}
\end{small}
\end{center}
\vskip -0.1in
\end{table}

\begin{table}[ht]
\caption{CTR Lift of new Ads in online experiment}
\label{newads}
\vskip 0.15in
\begin{center}
\begin{small}
\begin{sc}
\begin{tabular}{lcccr}
\hline
\abovespace\belowspace
Model & Lift & Percentage\\
\hline
DNN(Batch)    &  BaseLine & 11.86\% \\
FM-MLP(Online)   & +22.7\% & 16.72\% \\
FFM-MLP(Online)  & +25.6\% & 17.02\% \\
\hline
\end{tabular}
\end{sc}
\end{small}
\end{center}
\vskip -0.1in
\end{table}

\section{Notes}
\label{Notes}

CTR is a systematic project. In this section, we introduce some details in practical application.

\subsection{Negative Sampling}
Typical CTRs are much lower than 50\%, which means that positive examples (clicks) are relatively rare. Thus, simple statistical calculations indicate that clicks are relatively more valuable in learning CTR estimates. We can take advantage of this to significantly reduce the training data size with minimal impact on accuracy. We remain all positive examples and subsampling negative examples with a sample rate $w$. Then the predicted CTR as p, the recalibrated CTR $q$ should be $q = p/(p+(1-p)/w)$. In general, we will select a suitable $w$ to make the ratio of positive and negative samples being about 0.1.

\subsection{Burning}
Online models need long day's data to get stable effect. In practice, we want to know the real effect of experiment such as a new model and a new feature as soon as possible. For this purpose, we need to burn a new model with last serveral day's data. In general, we use 15-30 day's data to burn a new model.Because the complexity of deep model, we propose training the first several day's data with a single thread in burning and then with muti multiple threads. Otherwise the effect of new model is often unstable.

\subsection{Prior}
Because our model can be online learning, so the prior is not very important. The prior mean can be set to zero and prior variance can be set to 0.01 because more accurate and stable effect can be got by using smaller prior variance.

\subsection{Weight Decay}
If the model has run a long time, some variances of weights would converge towards zero and learning would come to a halt. Like AdPredictor and MatchBox, we make these variances converging back to the prior variance.

{\setlength\abovedisplayskip{1pt}
\setlength\belowdisplayskip{1pt}
\begin{equation}
\begin{split}
v' = \frac{vv_{prior}}{(1-\varepsilon)v_{prior}+\varepsilon v }
\end{split}
\end{equation}}

where $v$ is the current variance and $v_{prior}$ is the prior variance.

\subsection{Reinforcement}
Though our models only need to one pass of data, we can use more data or learn same data with multiple times for better performance. In fact, we test two methods to improvement effect but keep realtime update.
\begin{itemize}
\item {\bf Multiple Data :} We can use two or more sets of training workers with same config. These sets training the data with same positive examples and different negative examples because of negative sampling. The then get and update parameters with a same parameter server.
\item {\bf Batch Training :} We only use one set of training workers. But in every minibatch data, we run multiple ADF or EP on the data.
\end{itemize}
In our online experiments, both above methods can lead to better performance. However, the two methods is less safe when encountering abnormal data. So it is greater challenges to daily operations.

\section{CONCLUSION}
\label{submission}

We describe a parallel bayesian online deep learning framework used for click through rate prediction in tencent advertising system. And in the framework, we introduce several novel online deep probit regression model and get better performance than other known models in online experiments. Next we should explore more valuable online models in this framework. On the other hand, we would try more bayesian optimization methods to get better performance and more stable effect.

\nocite{fm}
\nocite{libfm}
\nocite{ffm}
\nocite{wd}
\nocite{ftrl}
\nocite{adpredictor}
\nocite{pnn}
\nocite{matchbox}
\nocite{ep}
\nocite{svb}
\nocite{pbp}
\nocite{cnnctr}
\nocite{rnnctr}
\nocite{cnn}
\nocite{rnn}
\nocite{gbdtlr}
\nocite{nnctr}
\nocite{ctrnad}
\nocite{spad}
\nocite{owlqn}

\bibliography{ctr}
\bibliographystyle{icml2013}

\end{document}